\documentclass[10pt, a4paper]{article}

\usepackage{lrec-coling2024}
\usepackage{booktabs}
\usepackage{CJKutf8}
\usepackage{colortbl}
\usepackage{amssymb}

\definecolor{Gray}{gray}{0.9}
\definecolor{LightCyan}{rgb}{0.88,1,1}
\newcolumntype{a}{>{\columncolor{Gray}}c}
\newcolumntype{b}{>{\columncolor{white}}c}

\title{JDocQA: Japanese Document Question Answering \\
Dataset for Generative Language Models}

\name{Eri Onami${}^{1,2}$, Shuhei Kurita${}^{2}$, Taiki Miyanishi${}^{3}$, Taro Watanabe${}^{1}$} 

\address{${}^{1}$Nara Institute of Science and Technology, ${}^{2}$RIKEN, ${}^{3}$ATR \\
         \{onami.eri.ob6, taro\}@is.naist.jp, shuhei.kurita@riken.jp, miyanishi@atr.jp}

\abstract{
Document question answering is a task of question answering on given documents such as reports, slides, pamphlets, and websites, and it is a truly demanding task as paper and electronic forms of documents are so common in our society.
This is known as a quite challenging task because it requires not only text understanding but also understanding of figures and tables, and hence visual question answering (VQA) methods are often examined in addition to textual approaches.
We introduce Japanese Document Question Answering (JDocQA), a large-scale document-based QA dataset, essentially requiring both visual and textual information to answer questions, which comprises 5,504 documents in PDF format and annotated 11,600 question-and-answer instances in Japanese.
Each QA instance includes references to the document pages and bounding boxes for the answer clues.
We incorporate multiple categories of questions and \textit{unanswerable} questions from the document for realistic question-answering applications.
We empirically evaluate the effectiveness of our dataset with text-based large language models (LLMs) and multimodal models. Incorporating \textit{unanswerable} questions in finetuning may contribute to harnessing the so-called hallucination generation.
\\ \newline 
}

\begin{document}

\maketitleabstract

\section{Introduction}

A thorough understanding of documents that are composed of both texts and graphical elements such as slides, reports, webpages, and pamphlets is essential for intelligent agents that process multimedia documents and answer some questions on such documents.
Document visual understandings have been studied to achieve joint understandings of textual and visual elements in such documents or images, including bookcovers~\cite{mathew2021docvqa}, scene images with characters~\cite{singh2019textvqa}, webpages~\cite{VisualMRC2021}, tables~\cite{Smock_2022_CVPR} and slides~\cite{SlideVQA2023}. 
These datasets have received significant attention as documents are a common form in various industrial, public, and private sectors in the English domain.
It is also notable that the document visual question answering tasks are still quite difficult despite its significance in industries because they heavily rely on both textual and visual modalities as the documents often include complex visual alignments of texts on figures, charts and illustrations. Especially in document question answering, models are required to connect multiple modalities to figure out answers.
There are quite limited datasets in which both visual and textual information is required to answer questions on documents.
It is also a problem that despite the significance of these tasks, the primary focus of these datasets is limited to the English domain and dataset constructions on other languages are still limited.
As a document question-answering task, Japanese documents have several characteristics compared to English documents.
One of the major difficulties in Japanese document processing lies in the two official writing styles in Japanese: one is a left-to-right horizontal style and the other is an upside-to-bottom vertical style, which requires both writing style comprehension in the dataset.

There has been significant progress in the generative large language models (LLMs) and multimodal models these days. GPT-4~\citep{OpenAI2023GPT4TR} allowing zero-shot applications in both language related and even in multimodal tasks.
InstructBLIP~\cite{dai2023instructblip} takes both textual and image inputs and generates texts such as image captions or visual question answering following textual prompts.
The success of LLMs also triggers the competitive development of several publicly-available LLMs in Japanese.
Instruction tuning of LLM can improve its ability to adhere to certain domains or usage, rendering them more suitable for particular applications rather than maintaining its ability in a general understanding of language and limited expertise~\citep{mishra-etal-2022-cross,sanh2022multitask}.
While there have been numerous attempts to fine-tune instruction for highly technical and professional adaptation~\citep{NEURIPS2022_b1efde53,wei2022finetuned}, there is still a lack of adequately prepared high-quality visual question and answering datasets that can be used for generative language model-based question answering and particularly developed outside of the English domain.

\begin{figure*}[t]
    \centering
    \includegraphics[width=15.5cm]{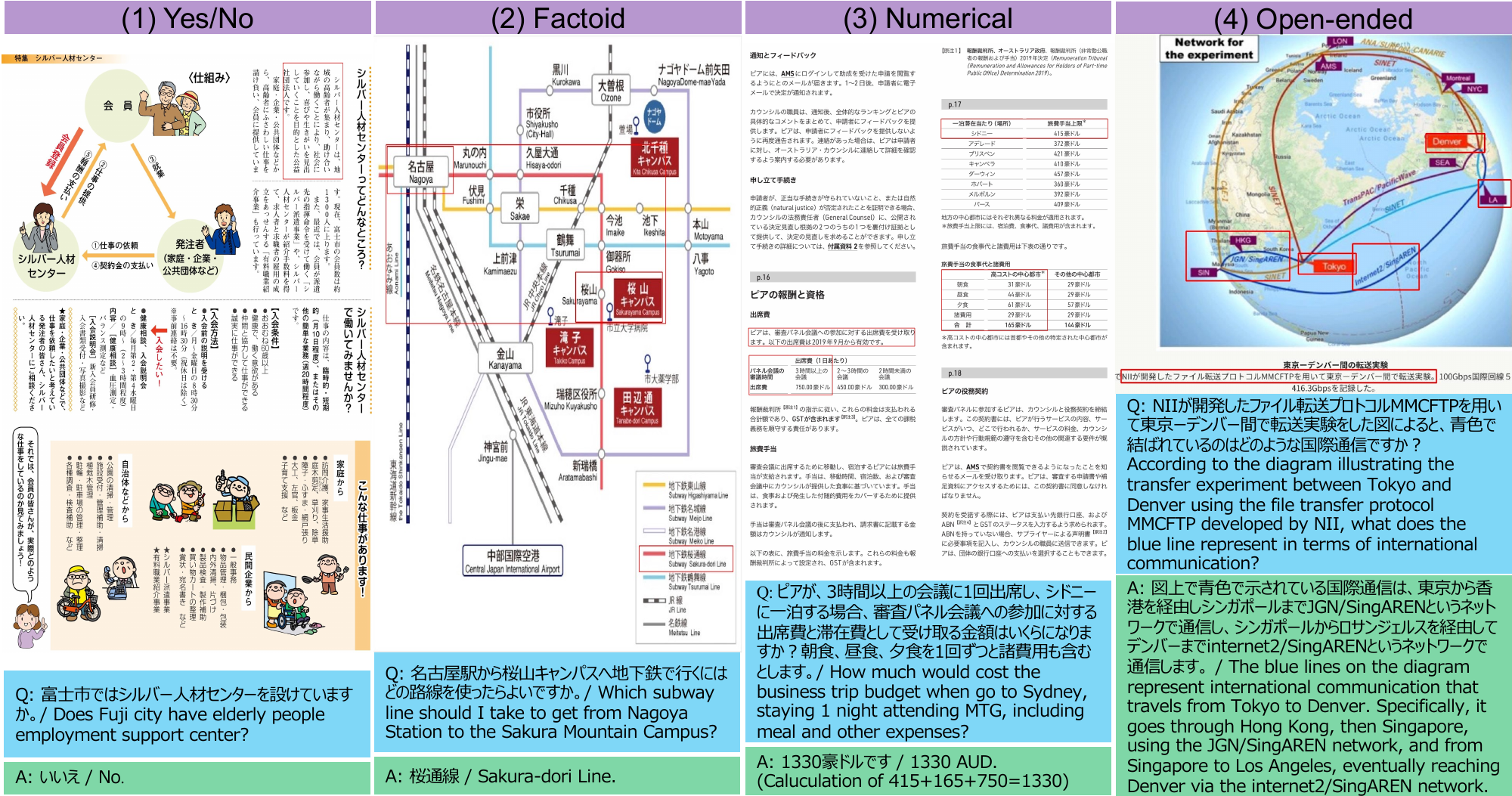}
    \caption{JDocQA sample question answering instances in four question categories with the annotated bounding boxes of the supporting facts in red color.}
    \label{fig:pdfqa_ano_ex}
\end{figure*}

To address the demand for a large-scale and fully annotated Japanese document question answering dataset, we introduce a JDocQA dataset by collecting Japanese documents in PDF styles from open-access sources including multiple formats of documents: slides, reports, websites and pamphlets and manually annotating question-answer pairs on them.
JDocQA consists of 11,600 question and answer pairs on the collected 5,504 documents as references for answering the question, four different question categories and 1,000 multi-page questions.
Each question is designed to refer to both textual and visual components such as tables or figures by annotators.
We also introduce \textit{unanswerable} questions: questions that have no answer clues in the referenced documents.
In experiments, we first present the effectiveness of finetuning LLM with our dataset.
We also suggest that incorporating these \textit{unanswerable} questions can contribute to mitigating \textit{hallucination}, which is often observed during the generation by LLMs.

\section{Related Work}
\paragraph{Multimodal question answering datasets.}
Visual question answering is a task of question-answering given visual contexts such as images following textual queries~\citep{NIPS2014_d516b136,VQA_2015}.
Earlier VQA studies have been not limited to images alone but cover various forms of media such as textbook~\citep{Kembhavi_2017_CVPR}, 
recipe~\citep{yagcioglu-etal-2018-recipeqa}, 
comic book~\citep{Iyyer2016TheAM},
movie~\citep{Tapaswi2016_MovieQA}.
Among them, a document VQA, which is a task designed for text embedded in real-world images, has attracted a lot of attention toward a comprehensive understanding of documents from both the visual and textual sides. 
Currently, some useful document VQA datasets have been published, such as OCR-VQA~\citep{mishraICDAR19}, TextVQA~\citep{singh2019textvqa}, and DocVQA~\citep{mathew2021docvqa}, VisualMRC~\citep{VisualMRC2021}, WebSRC~\citep{chen-etal-2021-websrc}, and InfographicVQA~\citep{InfographicVQA}.
Most of these studies concentrate on the single-image VQA where each question-answer pair has a single relevant image that always include sufficient information for question-answering.
Unlike the single-image VQA, the ability to comprehend multiple pages or charts to answer questions is more practical for understanding the slides and documents people read in the daily work.
To tackle such multi-image VQA, MultiModalQA~\citep{TalmorYCLWAIHB21}, MP-DocVQA~\citep{Tito2022HierarchicalMT} and SlideVQA~\citep{SlideVQA2023} concentrate on the multi-hop and numerical reasoning while considering multimodal context similar to previous works~\cite{rajpurkar-etal-2016-squad, yang-etal-2018-hotpotqa, dua-etal-2019-drop}.
It is also notable that in document question answering \citet{udop} proposed Universal Document Processing (UDOP), unifying vision, text, and layout of the input document through vision-text-layout Transformer.

\begin{table*}[t]
\centering
\scriptsize\begin{tabular}{lllllllll}
\hline
\textbf{Category} & \textbf{Documents} & \textbf{QA} & \textbf{(1) Yes/No} & \textbf{(2) Factoid} & \textbf{(3) Numerical} & \textbf{(4) Open-ended} & \textbf{Multi-page}& \textbf{Unanswerable}\\
\hline
Pamphlet & 1,715 & 4,025 & 605 & 748 & 660 & 2,012 &  46 & 671\\
Slide & 1,640 & 3,276 & 545 & 593 & 507 & 1,631 & 448 & 449\\
Report & 2,086 & 4,167 & 703 & 687 & 693 & 2,084 &  506 & 668\\
Website & 67 & 132 & 2 & 24 & 6 & 100 & 0 & 0 \\
\hline
Total & 5,504 & 11,600 & 1,855 & 2,052 & 1,866 & 5,827 & 1,000 & 1,788\\
\hline
\end{tabular}
\caption{\label{table:dataset_statistics_1}
Number of document styles and question-answer pairs by the four question categories, multi-page and unanswerable questions. 
}
\end{table*}

\begin{table}[t]
\centering
\scriptsize\begin{tabular}{lllllllll}
\hline
\textbf{Category} & \textbf{(1) Y/N} & \textbf{(2) Fact.} & \textbf{(3) 
 Num.} & \textbf{(4) Open.}\\
\hline
Context & 963.81 & 1036.63 & 1020.04 & 1017.25\\
Question & 67.75 & 61.26 & 60.36 & 65.44\\
Answer & 3.77 & 16.01 & 8.22 & 65.97\\
\hline
\end{tabular}
\caption{Average character length.}
\label{table:dataset_statistics_3}
\end{table}

\begin{table*}[t]
\centering
\scriptsize\begin{tabular}{llllllll}
\hline
\textbf{Dataset}  & \textbf{\#Questions} & \textbf{\#Images} & \textbf{\#BBoxes} & \textbf{Language} & \textbf{Multihop}\\
\hline
OCR-VQA~\citep{mishraICDAR19}  & 1002k & 207k & - & English & -\\
DocVQA~\citep{mathew2021docvqa} & 50k & 12k & - & English & - \\
InfographicVQA~\citep{InfographicVQA}  & 5.9k & 30k & - & English & -\\
MP-DocVQA~\citep{Tito2022HierarchicalMT}  & 46k & 48k & - & English & \checkmark\\
SlideVQA~\citep{SlideVQA2023}  & 14.5k & 52k & 890k & English & \checkmark\\
JDocQA (Ours) & 11.6k & 268k & 11k & Japanese & \checkmark\\
\hline
\end{tabular}
\caption{The comparison of the document question answering datasets.}
\label{table:dataset_statistics_4}
\end{table*}

\paragraph{Text-based question answering in Japanese.}
Some related tasks of text-based question answering for Japanese have been studied~\citep{miyazaki-shimizu-2016-cross,yanaka-mineshima-2022-compositional,takahashi-etal-2019-machine,kurihara-etal-2022-jglue}.
\citet{miyazaki-shimizu-2016-cross} created a Japanese image captioning dataset, which is the Japanese version of the MS-COCO captions dataset, and demonstrated that using both bilingual datasets outperforms using monolingual ones.
\citet{yanaka-mineshima-2022-compositional} introduced a Japanese textual entailment dataset and highlighted that many existing models that have focused on English do not adequately account for Japanese language characteristics.
\citet{takahashi-etal-2019-machine} proposed a QA dataset based on Japanese blogs related to driving, with the aim of creating a model that can understand the meaning of sentences or texts in Japanese.
It is also notable that \citet{miyao-kawazoe-2013-university} and relevant Todai Robot Project\footnote{\url{https://21robot.org/index-e.html}} arranged the dataset of Japanese University entrance-exams. In math and physics subjects, their dataset includes limited multimodal contents in DTD file format, although it doesn't cover general domains as of JDocQA.
Among the datasets related to the Japanese language, JGLUE~\citep{kurihara-etal-2022-jglue} is similar to our work in terms of the aim of the datasets.
JGLUE is a large-scale natural language understanding (NLU) benchmark purposed for the evaluation of LLM.
It includes various tasks, such as text classification, sentence pair classification, and QA to assess Japanese comprehension.
In contrast to JGLUE, our dataset offers a diverse range of question types which can be useful for instruction tuning and contains both Japanese text and image data, which can be used for multimodal models.
Another key difference with JGLUE is that JDocQA incorporates unanswerable questions to help suppress hallucinations.

\section{Dataset}

\subsection{Task Overview and Formulation}
\label{sec:taskoverview}
We consider generative question answering where a model generates a textual answer following the document context and textual question. For realistic applications of a wide range of user questions for documents, we prepare four categories of questions: (1) \textbf{yes/no}, (2) \textbf{factoid}, (3) \textbf{numerical}, and (4) \textbf{open-ended}.
In yes/no questions, answers are ``yes'' or ``no.''
In factoid questions, answers are some facts, such as named entities, that typically appear in the given documents.
In numerical questions, answers are numeric values, often including some numerals (some units, e.g., \textit{km} or Japanese numerals such as ``\begin{CJK}{UTF8}{ipxm}個\end{CJK} (\textit{objects})'' and ``\begin{CJK}{UTF8}{ipxm}人\end{CJK} (\textit{persons})''). These numeric values are written in the documents or are calculated from other numbers in the documents.
In open-ended questions, free-form responses are required. For such questions, we aim to assess complex comprehension abilities, such as the ability to form opinions or brief explanations based on the provided contexts and questions.
Figure~\ref{fig:pdfqa_ano_ex} presents samples of these four categories of questions.
All examples include diverse images and question types related to some Japanese documents collected.
We also include \textit{unanswerable} questions for each question category.

In the realistic applications of the question answering, no answers can be found in the referenced document.
Therefore, it is expected that the correct responses for such questions are ``not mentioned in the text.''
The prediction of the \textit{unanswerable} questions is not addressed in previous Japanese question answering datasets such as \citet{kurihara-etal-2022-jglue}.

\begin{figure}[t]
    \centering
    \includegraphics[width=8.7cm]{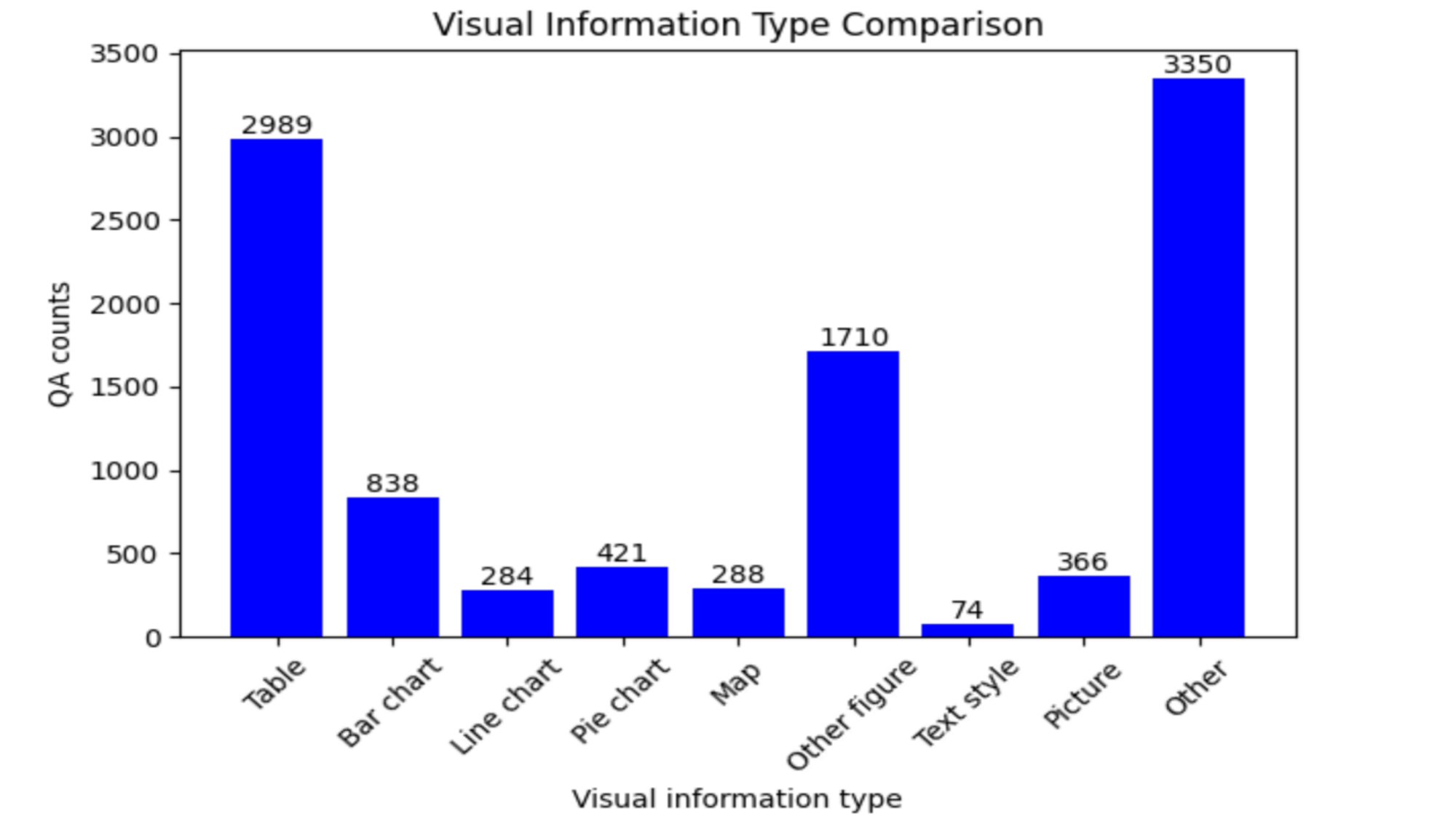}
    \caption{The number of visual information categories.}
    \label{fig:category}
\end{figure}

\subsection{Dataset Statistics}
JDocQA dataset comprises 5,504 files and 11,600 question-and-answer pairs in Japanese.
The statistics of categorized question types are as follows: (1) yes/no questions: 1,855, (2) factoid questions: 2,052, (3) numerical questions: 1,866, (4) open-ended questions: 5,827. 
Additionally, 1,788 questions require referencing multiple pages to answer, and in 1,000 questions the correct answer is not mentioned in the text, as shown in Table~\ref{table:dataset_statistics_1}.
Some PDF documents contain both slide and report formats within the same documents. 
For such documents, we count them in both categories of the slide and report formats when we calculate the total number of the documents\footnote{The number of total documents is not the sum of the number of the subcategories in Table~\ref{table:dataset_statistics_1} due to this counting.}.
Table~\ref{table:dataset_statistics_3} represents the average length of the context, question, and answer in our dataset, and Figure~\ref{fig:category} shows the category of the visual information referenced by question or answer in our dataset.
The comparison of document question answering datasets are shown in Table~\ref{table:dataset_statistics_4}.

\subsection{Dataset Creation}
\label{sec:dataset collection}
\begin{figure}[t]
    \centering
    \includegraphics[width=7.5cm]{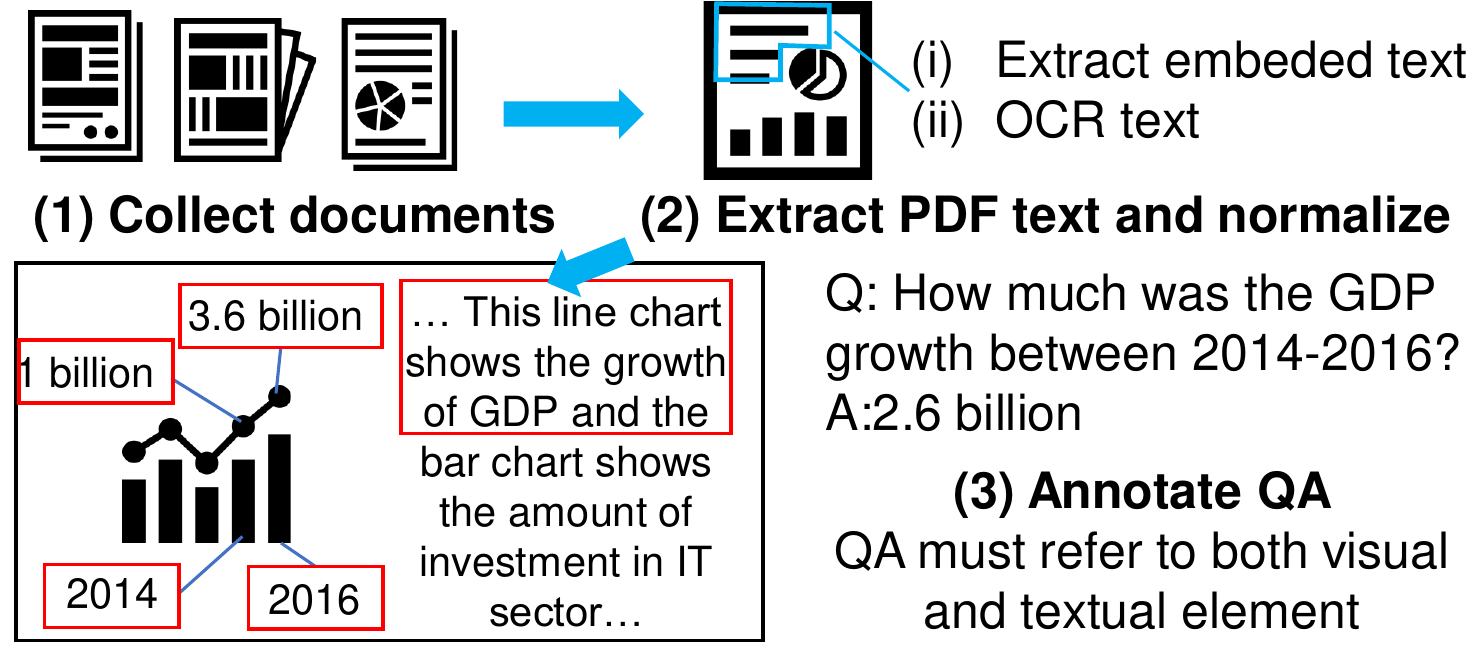}
    \caption{Annotation process.}
    \label{fig:annotation_process}
\end{figure}

The overall dataset creation and annotation process is presented in Figure~\ref{fig:annotation_process}.
\paragraph{PDF collection.} 
We gather public documents, such as, municipality pamphlets and websites, that are created by Japanese governmental agencies or local governments.
We manually collected PDF documents from open-access resources such as Japanese National Diet Library (NDL)'s digital collection, web archive projects (WARP)\footnote{\url{https://warp.ndl.go.jp/}} and websites of Japanese government ministries.
We manually gathered documents such as reports, pamphlets or websites that are published by public or quasi-public sectors, such as local governments or public universities through WARP. We also gather Japanese ministry documents such as slides and reports from their websites following the government agencies' policies.
Those documents cover a wide range of topics, for instance, economic policies, education policies, labor issues, health and hygiene, agriculture, forestry, fisheries, culture and arts, history, related to governmental policy or policy guidelines, as well as the everyday affairs of local governments.
These documents also include visual elements such as figures, tables, charts, pictures, or mandala charts, complex figures with a combination of texts and objects typically seen in the Japanese public administrative sector's official document.
We classify these documents into four categories, namely, \textbf{pamphlet}, \textbf{slide}, \textbf{report}, and \textbf{website} considering the form of the documents.

\paragraph{Text Extraction \& Normalization.}
\label{sec:context normalization}
We extracted texts from PDF documents with PyPDF2\footnote{We also examined PyMuPDF. However, the quality of the extracted texts was not changed greatly.}.
We also notice that some PDF documents are probably created from paper scans, and we cannot extract embedded texts from such documents.
Therefore, we extracted texts from the document page images by OCR (Optical Character Recognition) as an alternative source. After the text extraction or OCR, we removed mistakenly recognized symbols and emojis, or duplicated characters from texts when the same character continuously and repeatedly appeared more than five times.

\paragraph{Annotation Procedure.} 
We ask 43 annotators in total for the question-answering pairs annotation on documents.
As documents include rich textual and visual elements (e.g., graphs, charts, maps, illustrations, and a mix of vertical and horizontal written text), we made question answer pairs that are related to both textual and visual information.
We ask annotators to write up two to four question-answer annotations in each document.
We also ask not to use any AI-tools such as OpenAI ChatGPT during the annotation process.
Each question is accompanied with the supporting facts as marked in red in Figure~\ref{fig:pdfqa_ano_ex} and Figure~\ref{fig:annotation_process}.
We classify a subset of questions that have multiple supporting facts in multiple pages as \textit{multi-page} questions.
Multi-page questions are considerably difficult from their single-page counterparts.
For unanswerable questions, we ask annotators to write questions that lack supporting facts in the documents, making them impossible to answer based on the given documents.

\paragraph{Visual inputs and bounding boxes.}
\label{sec:visual inputs}
We prepared three types of images for visual inputs for multimodal models. The first type of images are those of the whole page of the documents including the annotated question answering pairs. The second type of images are those cropped by bounding boxes on which annotators based their answers such as tables or figures of the pages. When multiple bounding boxes are annotated to a single question-answer pair, multiple cropped images are combined together into a single image here. The third type of images are blank (white) images that are used for ablation studies.

\section{Experiments}

\subsection{Question-Answering Task}
Our dataset aims to evaluate the question answering ability following the document contexts, including textual and visual information, and questions via open-ended text generation.
As discussed in Sec.~\ref{sec:taskoverview}, our dataset consists of four forms of questions: yes/no\footnote{The chance rate of yes/no questions including \textit{unanswerable} is 61.57 when the model always marks ``yes.''}, factoid, numerical, and open-ended.
All of these four category questions include \textit{unanswerable} questions that are not answerable solely from the given document file.
Models are expected to generate answers in open-ended text generation in any of these question types.
Textual model inputs, or simply \textit{prompts}, consist of the embedded texts or OCR results of documents as described in Sec.~\ref{sec:context normalization} and the questions. We also include the answer-format guidelines such as ``please answer in Yes/No form'', ``please answer the fact that is referred to in the document'', ``please answer by numerical information from the document'' and ``please write the answer in open-ended format'' into the prompts depending on the four question categories.
For \textit{unanswerable} questions, we prepare a special answer for all question categories: ``\begin{CJK}{UTF8}{ipxm}本文中に記載がありません\end{CJK} (\textit{not mentioned in the text}).''

\subsection{Models}
We conduct experiments with both text-input models and multimodal models of text and vision inputs.
For model training, we use supervised finetuning. The best hyperparameters are searched with train and validation sets, then the model performance is evaluated with the best hyperparameters.

\paragraph{Models with text input.}
We adapted up to 13 billion (13B) model parameter scale Japanese large language models for experiments.
We experimented with the following representative Japanese models that take only textual inputs:
rinna japanese-gpt2-medium\footnote{\url{https://huggingface.co/rinna/japanese-gpt2-medium}}, 
japanese-gpt-4B-8k\footnote{\url{https://huggingface.co/rinna/bilingual-gpt-neox-4b-8k}},
rinna japanese-gpt-1B\footnote{\url{https://huggingface.co/rinna/japanese-gpt-1b}},
Cyberagent OpenCALM-7B\footnote{\url{https://huggingface.co/cyberagent/open-calm-7b}},
Matsuo-Lab weblab-10b\footnote{\url{https://huggingface.co/matsuo-lab/weblab-10b}}, 
PFNet PLaMo-13B\footnote{\url{ https://huggingface.co/pfnet/plamo-13b}}, 
Stability AI Japanese-StableLM-Base-Alpha-7B\footnote{\url{https://huggingface.co/stabilityai/japanese-stablelm-base-alpha-7b}}, and
Stability AI Japanese-StableLM-Instruct-Alpha-7B\footnote{\url{https://huggingface.co/stabilityai/japanese-stablelm-instruct-alpha-7b}}.
We also include multilingual large language model of Llama-2-7B~\footnote{\url{https://huggingface.co/meta-llama/Llama-2-7b-hf}}.
We trained and evaluated models with 1024 token length for fair comparisons and computational efficiency except rinna japanese-gpt-4B-8k which is trained with 8192 tokens at most.
For analyses with longer token lengths, we train rinna japanese-gpt-4B-8k model with 2048, 4096, and 8192 tokens.


\begin{table*}[t]
\centering
\scriptsize\begin{tabular}{laccccacccc}
\hline
    &\multicolumn{5}{c}{Validation set}  & \multicolumn{5}{c}{Test set} \\
    \cmidrule(lr){2-6} \cmidrule(lr){7-11} 
\textbf{Model} & \textbf{Avg.} & \textbf{(1) Y/N} & \textbf{(2) Fact.} & \textbf{(3) Num.} & \textbf{(4) Open.} & \textbf{Avg.}& \textbf{(1) Y/N}& \textbf{(2) Fact.}& \textbf{(3) Num.} & \textbf{(4) Open.} \\
\hline
\multicolumn{11}{l}{\textit{Evaluated with all instances.}} \\
gpt-3.5-turbo-16k                   & 19.86 & 47.89 & 7.85 & 7.97 & 15.75    & 20.62 & 50.29 & 7.44 & 11.11 & 13.64 \\
gpt-4                               & 17.96 & 34.73 & 9.42 & 8.51 & 19.17    & 19.47 & 43.19 & 6.51 & 11.11 & 17.07 \\
\hline
\multicolumn{11}{l}{\textit{Evaluated without ``unanswerable.''}} \\
gpt-3.5-turbo-16k                   & 22.72 & 57.23 &  8.82 & 9.20 & 15.63    & 23.07 & 58.21 & 8.08 & 12.5 & 13.49\\
gpt-4                               & 20.90 & 41.50 & 10.58 & 9.81 & 21.72    & 22.03 & 50.00 & 7.07 & 12.5 & 18.57 \\
\hline
\multicolumn{11}{l}{\textit{Models trained with all training instances and evaluated all instances.}} \\
rinna gpt2-medium-336M              & 21.33 & 63.15 &  7.32 &  4.78 & 17.51  &  19.41 & 62.13 &  4.65 &  8.18 & 15.99 \\
rinna gpt-1B                        & 23.79 & 58.42 & 10.99 &  6.38 & 22.27  &  20.46 & 59.76 &  5.58 &  8.77 & 18.13 \\
rinna bi-4B-8k (8192 tok.)                & 26.35 & 55.26 & 14.65 & 13.29 & 24.93  &  23.02 & 62.13 &  8.83 & 11.11 & 20.57 \\
OpenCALM-7B                         & 21.65 & 47.36 & 14.65 &  5.31 & 20.81  &  18.33 & 43.78 & 11.62 &  9.94 & 16.03 \\
weblab-10B                          & 19.20 & 46.31 &  9.42 &  7.97 & 17.13  &  16.94 & 47.92 & 10.23 &  8.18 & 13.24 \\
PLaMo-13B                           & 25.79 & 55.26 & 15.18 & 15.42 & 22.92  &  20.33 & 53.84 & 10.69 &  7.01 & 18.21 \\
StableLM Base-Al.-7B              & 32.92 & 67.89 & 21.98 & 18.61 & 29.62  &  29.71 & 70.41 & 15.81 & 22.22 & 25.51 \\
StableLM Inst.-Al.-7B          & 33.80 & 67.36 & 20.94 & 20.21 & 31.39  &  29.56 & 72.78 & 16.27 & 21.05 & 24.75 \\
Llama2-7B                              & 30.29 & 60.00 & 20.41 & 15.42 & 28.59  &  27.01 & 61.53 & 17.20 & 18.71 & 23.29 \\
\hline
\multicolumn{11}{l}{\textit{Models trained with all training instances while evaluated without ``unanswerable.''}} \\
rinna gpt2-medium-336M              & 22.22 & 69.81 &  4.70 &  3.68 & 18.71  &  19.61 & 65.75 &  4.54 &  6.57 & 16.31 \\
rinna gpt-1B                        & 22.84 & 62.26 &  7.05 &  5.52 & 21.14  &  20.18 & 64.38 &  4.04 &  8.55 & 17.40 \\
rinna bi-4B-8k (8192 tok.)                       & 24.09 & 52.83 & 12.94 & 10.42 & 23.10  &  21.17 & 62.32 &  6.56 &  7.89 & 19.11 \\
OpenCALM-7B                         & 20.75 & 50.31 & 12.35 &  4.90 & 19.20  &  17.53 & 46.57 &  9.59 &  8.55 & 15.10 \\
weblab-10B                          & 17.74 & 45.91 &  8.82 &  6.13 & 15.34  &  16.20 & 50.68 &  9.09 &  6.57 & 12.17 \\
PLaMo-13B                           & 22.66 & 54.08 & 10.58 &  9.81 & 20.76  &  18.35 & 50.68 &  8.58 &  5.26 & 16.87 \\
StableLM Base-Al.-7B              & 31.55 & 70.44 & 18.82 & 15.95 & 28.24  &  28.33 & 71.91 & 12.62 & 21.05 & 24.32 \\
StableLM Inst.-Al.-7B          & 31.74 & 69.81 & 17.05 & 15.95 & 29.55  &  28.66 & 76.02 & 12.62 & 19.07 & 24.40 \\
Llama2-7B                              & 28.78 & 62.26 & 17.05 & 14.72 & 26.44  &  25.70 & 65.06 & 12.62 & 18.42 & 21.87 \\
\hline
\multicolumn{11}{l}{\textit{Models trained without ``unanswerable'' while evaluated with all instances.}} \\
rinna gpt2-medium-336M              & 15.91 & 52.10 &  3.14 &  3.72 & 12.11  &  17.81 & 63.31 &  4.18 &  5.26 & 13.60 \\
rinna gpt-1B                        & 17.66 & 45.26 &  6.80 &  2.65 & 17.04  &  17.59 & 51.47 &  6.04 &  5.26 & 15.76 \\
rinna bi-4B-8k (8192 tok.)                       & 23.44 & 57.89 & 12.04 &  9.57 & 20.33  &  23.01 & 69.23 &  7.90 &  9.94 & 19.27 \\
OpenCALM-7B                         & 18.94 & 42.63 &  8.90 &  3.72 & 19.44  &  16.95 & 42.01 &  7.44 &  6.43 & 16.33 \\
weblab-10B                          & 20.43 & 50.00 &  9.42 &  6.91 & 18.70  &  17.96 & 52.07 &  6.51 &  8.18 & 15.34 \\
PLaMo-13B                           & 22.04 & 60.00 &  7.85 &  9.57 & 18.23  &  21.11 & 64.49 &  8.83 & 11.11 & 16.31 \\
StableLM Base-Al.-7B              & 27.02 & 63.68 & 14.65 & 12.76 & 23.61  &  25.68 & 68.63 & 12.09 & 17.54 & 20.94 \\
StableLM Inst.-Al.-7B          & 27.22 & 61.57 & 15.70 & 14.36 & 23.84  &  26.25 & 70.41 & 15.34 & 16.37 & 20.74 \\
Llama2-7B                              & 30.15 & 63.68 & 18.32 & 13.82 & 28.30  &  28.25 & 73.96 & 11.62 & 16.95 & 24.68 \\
\hline
\multicolumn{11}{l}{\textit{Models trained without ``unanswerable'' and evaluated without ``unanswerable'' instances.}} \\
rinna gpt2-medium-336M              & 18.75 & 62.26 &  3.52 &  4.29 & 14.33  &  20.12 & 73.28 &  4.54 &  5.92 & 15.41 \\
rinna gpt-1B                        & 21.15 & 54.08 &  7.64 &  3.06 & 21.17  &  20.02 & 59.58 &  6.56 &  5.92 & 18.20 \\
rinna bi-4B-8k (8192 tok.)                       & 27.63 & 69.18 & 13.52 & 11.04 & 24.26  &  25.95 & 80.13 &  8.58 & 11.18 & 21.79 \\
OpenCALM-7B                         & 20.55 & 50.94 & 10.00 &  4.29 & 19.67  &  17.97 & 48.63 &  8.08 &  7.23 & 16.32 \\
weblab-10B                          & 22.25 & 59.74 & 10.58 &  7.97 & 18.55  &  19.06 & 60.27 &  7.07 &  9.21 & 15.06 \\
PLaMo-13B                           & 26.48 & 71.69 &  8.82 & 11.04 & 22.74  &  24.29 & 74.65 &  9.59 & 12.50 & 19.34 \\
StableLM Base-Al.-7B              & 32.30 & 76.10 & 16.47 & 14.72 & 29.15  &  29.21 & 79.45 & 13.13 & 19.73 & 24.14 \\
StableLM Inst.-Al.-7B          & 32.41 & 73.58 & 17.64 & 16.56 & 29.16  &  29.75 & 81.50 & 16.66 & 18.42 & 23.69 \\
Llama2-7B                              & 33.11 & 76.10 & 20.58 & 15.95 & 28.84  &  30.57 & 85.61 & 12.62 & 19.07 & 25.47 \\
\hline
\end{tabular}
\caption{
Results of all finetuned models and OpenAI GPT zeroshot.
\textbf{Avg.} is weighted average of scores.
}
\label{table:generation_performance}
\end{table*}

\paragraph{Models with multimodal input.}
The purpose of JDocQA is to analyze documents with textual and visual perceptions.
To assess the impact of using both images and text on the JDocQA dataset, we applied multimodal models that take inputs from both images and texts.
We used Stability AI Japanese-StableLM-Instruct-Alpha-7B\footnote{\url{https://huggingface.co/stabilityai/japanese-stablelm-instruct-alpha-7b}}, a Japanese version of InstructBLIP~\cite{dai2023instructblip, li2023blip, li2023blip2} for this purpose, as they are applicable to Japanese text and image inputs.
We trained and evaluated this model with 512 token lengths following its max capacity.
We develop three different models for with three different visual inputs as explained in Sec.~\ref{sec:visual inputs}.
The first model takes visual inputs of a blank image that is always the same white image of the 800x600 pixel size as ablation study.
The second model takes an image of a whole document page that are related to the question-answering in the annotation. These images are also scaled to 800-pixel width.
The third model take inputs of the images following the annotated supporting facts. Following the annotated bounding boxes, we crop the referenced regions of the page images, combine the bounding boxes and scale them for the model visual input.
As some questions have several annotated supporting facts to their answers, the combined image may contain more than one region of the annotated bounding boxes. 
All of these multimodal models also take textual prompts that are similar to the text-input models.

\paragraph{OpenAI GPT baselines.}
We also present OpenAI GPT performances as baselines. Here we use gpt-3.5-turbo-16k and gpt-4 models\footnote{Latest model at Oct. 9 2023.}. They take similar prompts to those of text-input models. However, as they are the zero-shot models for our task, we observed they are quite sensitive to the prompts. To improve their performance, we manually tune the prompts for OpenAI GPT models. We avoid finetuning OpenAI GPT models although finetuning them may greatly improve performances due to the following reasons. First, our purpose is to develop local models that work on limited computational resources, Second, the details of finetuning are unavailable for these models, and finally due to the API cost issues.

\subsection{Evaluation Methods}
For Yes/No, factoid, and numerical questions, we used the exact match metric after trimming trivial differences such as the presence or absence of punctuation marks or Japanese suffix phrases such as ``\begin{CJK}{UTF8}{ipxm}です\end{CJK} (\textit{is})'' by simple rules.
We have also examined the variations of exact match, such as the ratio of whether model prediction phrases are included in the correct answer phrases or not. However, we realize that this metric performs quite similarly to the exact match evaluation and the difference is typically less than 10 question-answer pairs in the validation set.
For open-ended questions, answers are typically long, e.g., the average length of answers is 65.97 characters in Japanese, and hence the exact match does not work for evaluation. Therefore, we used BLEU score\footnote{\url{https://github.com/mjpost/sacrebleu}} tokenized by MeCab\footnote{\url{https://pypi.org/project/mecab-python3/}} for automatic evaluation of open-ended questions.

\subsection{Experimental Settings}
\label{sec:experimental_settings}
Our dataset includes \textit{unanswerable} questions in all question categories. While it is expected that finetuning models with unanswerable instances may harness models to surpass illusion answers known as hallucinations, detecting unanswerable questions is also notoriously difficult as of \citet{squad2.0}.
Therefore we prepare two types of models for all base models experimented: models finetuned with all question answering pairs including unanswerable instances and models finetuned without unanswerable instances.
In the evaluation, we similarly prepare two separate validation and test sets: the standard validation and test sets that consists of all question-answering pairs, and the smaller validation and test sets where unanswerable questions are removed.

\subsection{Results}
Table~\ref{table:generation_performance} presents the performance of text-input models for all question types on valid and test splits.

\paragraph{Models trained with all instances.}
We compare the first and third blocks in Table~\ref{table:generation_performance}.
They are the results of all JDocQA instances by zero-shot and finetuned models that are trained with all JDocQA training instances including ``unanswerable questions''.
We realize that fine-tuned models outperformed gpt-3.5 and gpt-4 results especially when the model size is larger.
Next, we compare the second and fourth blocks in Table~\ref{table:generation_performance}.
They are the same models with previous block evaluations on the JDocQA without unanswerable questions.
We observe a similar tendency to all instances evaluations, suggesting that models finetuned with unanswerable instances perform similar performances in both answerable and unanswerable questions.
Among them, StableLM models perform best despite their parameter size of 7B.
The rinna bi-4B-8k model also performs well despite its parameter size. 
We attribute this   to its token length size of 8192 and will discuss this later.
We notice that (1) yes/no questions are relatively easy although they do not have much effect on the averaged score (Avg.) that mostly follows the most common question category of (4) open-ended.

\begin{table}[t]
\centering
\scriptsize\begin{tabular}{lacccc}
\hline
    & \multicolumn{5}{c}{Test set} \\
    \cmidrule(lr){2-6} 
\textbf{Model} & \textbf{Avg.} & \textbf{(1) Y/N}\hspace{-1em} & \textbf{(2) Fact.}\hspace{-1em} & \textbf{(3) Num.}\hspace{-1em} & \textbf{(4) Open.} \\
\hline
\multicolumn{6}{l}{\textit{Trained all and evaluated all.}} \\
InstBLIP (blank)         & 26.92 & 65.68 & 16.27 & 19.88 & 22.00 \\
InstBLIP (img)           & 27.44 & 68.63 & 15.34 & 19.88 & 22.50 \\
InstBLIP (bbox)          & 27.87 & 72.78 & 18.13 & 19.29 & 21.37 \\
\hline
\multicolumn{6}{l}{\textit{Trained all while evaluated w/o ``unanswerable.''}} \\
InstBLIP (blank)         & 25.12 & 65.75 & 10.60 & 17.10 & 21.68 \\
InstBLIP (img)           & 25.74 & 69.17 & 11.61 & 15.13 & 22.12 \\
InstBLIP (bbox)          & 27.99 & 78.76 & 14.14 & 17.76 & 22.16 \\
\hline
\multicolumn{6}{l}{\textit{Trained w/o ``unanswerable'' while evaluated all.}} \\
InstBLIP (blank)         & 23.13 & 66.27 & 12.55 & 11.69 & 18.21 \\
InstBLIP (img)           & 25.01 & 71.59 & 12.09 & 16.37 & 19.19 \\
InstBLIP (bbox)          & 29.00 & 78.10 & 14.88 & 19.29 & 23.19 \\
\hline
\multicolumn{6}{l}{\textit{Trained w/o ``unanswerable'' and evaluated w/o ``unanswerable.''}} \\
InstBLIP (blank)         & 26.45 & 76.71 & 13.63 & 13.15 & 21.26 \\
InstBLIP (img)           & 28.52 & 82.87 & 13.13 & 18.42 & 22.25 \\
InstBLIP (bbox)          & 27.79 & 80.13 & 11.61 & 16.44 & 22.71 \\
\hline
\end{tabular}
\caption{
Results of multimodal input models.
\textbf{Avg.} is weighted average.
}
\label{table:multimodal_model}
\end{table}

\begin{table}[t]
\centering
\scriptsize\begin{tabular}{lacccc}
\hline
    & \multicolumn{5}{c}{Test set} \\
    \cmidrule(lr){2-6} 
\textbf{Token length} & \textbf{Avg.} & \textbf{(1) Y/N}\hspace{-1em} & \textbf{(2) Fact.}\hspace{-1em} & \textbf{(3) Num.}\hspace{-1em} & \textbf{(4) Open.} \\
\hline
\multicolumn{6}{l}{\textit{Trained all and evaluated all.}} \\
2048 tokens          &  20.97 & 57.39 & 10.69 &  9.94 & 17.66 \\
4096 tokens          &  21.96 & 56.21 &  9.30 & 13.45 & 19.38 \\
8192 tokens          &  23.02 & 62.13 &  8.83 & 11.11 & 20.57 \\
\hline
\multicolumn{6}{l}{\textit{Trained w/o ``unanswerable'' and evaluated w/o ``unanswerable.''}} \\
2048 tokens         &  24.57 & 72.60 & 10.10 &  9.21 & 21.18 \\
4096 tokens         &  24.26 & 67.12 &  9.09 & 11.18 & 21.90 \\
8192 tokens         &  25.95 & 80.13 &  8.58 & 11.18 & 21.79 \\
\hline
\end{tabular}
\caption{
Results of rinna bi-4B-8k models with different token length.
\textbf{Avg.} is weighted average.
}
\label{table:different length}
\end{table}

\begin{table}[t]
\centering
\scriptsize\begin{tabular}{lcccc}
\hline
    & \multicolumn{4}{c}{Test set} \\
    \cmidrule(lr){2-5} 
\textbf{Model} &  \textbf{Pamphlet}& \textbf{Slide}& \textbf{Report} & \textbf{Website} \\
\hline
\multicolumn{5}{l}{\textit{Trained all and evaluated all.}} \\
rinna gpt2-med-336M\hspace{-3em}   &  18.62 & 16.32 & 14.57 &  3.09 \\
rinna gpt-1B                        &  16.66 & 15.10 & 15.08 &  4.72 \\
rinna bi-4B-8k (8192) \hspace{-1.5em} &  21.81 & 16.73 & 18.41 &  3.53 \\
OpenCALM-7B                         &  15.19 & 15.10 & 13.04 &  2.56 \\
weblab-10B                          &  14.70 & 17.55 & 13.04 &  2.69 \\
PLaMo-13B                           &  21.56 & 13.06 & 15.85 &  2.90 \\
Base-Al.-7B                &  26.96 & 20.40 & 24.04 &  4.64 \\
Inst-Al.-7B                &  27.69 & 23.26 & 23.01 &  5.71 \\
\hline
InstBLIP-Al (blank)        & 25.49 & 22.04 & 21.48 &  3.79 \\
InstBLIP-Al (img)          & 25.00 & 21.63 & 23.78 &  3.75 \\
InstBLIP-Al (bbox)         & 25.00 & 22.04 & 22.50 &  4.20 \\
\hline
\end{tabular}
\caption{
Detailed file-type result. ``\textbf{Website}'' is included only in test set as an out-of-domain set.
}
\label{table:detailed_file_type}
\end{table}

\paragraph{Models trained without unanswerable.}
As explained in Sec.~\ref{sec:experimental_settings}, we also prepared models finetuned without \textit{unanswerable} questions in the training instances.
We present the evaluation results including and excluding \textit{unanswerable} questions in the fifth and sixth blocks in Table~\ref{table:generation_performance}. 
It is quite interesting when we compare the third and fifth block results in Table~\ref{table:generation_performance} as they share the same evaluation set including \textit{unanswerable} questions while the models are finetuned with and without the \textit{unanswerable} instances.
Comparing the third and fifth block results, we notice almost all models finetuned with \textit{unanswerable} questions perform better than their \textit{answerable}-only finetuned counterparts in the averaged scores. 
Exceptions are the OpenCALM-7B, weblab-10B and Llama2-7B models, which we will discuss in the next paragraph.
We attribute this is due to the concept of \textit{hallucination,} where models generate answers that do not appear in context texts. 
We will present an example of this in the qualitative analysis paragraph.
Doping \textit{unanswerable} instances may contribute to harnessing \textit{hallucination} in this experimental comparison.

\paragraph{OpenCALM-7B and weblab-10B do not predict questions as \textit{unanswerable} so much.}
In the third block of Table~\ref{table:generation_performance}, we notice interesting phenomena: OpenCALM-7B and weblab-10B do not perform well despite their parameter size. When we closely check these models' outputs, we realize these models, finetuned with all instances, predict ``\begin{CJK}{UTF8}{ipxm}本文中に記載がありません\end{CJK} (\textit{not mentioned in the text})'' much less than other models. OpenCALM-7B and weblab-10B predicts 11.9\% and 13.7\% of all instances are \textit{unanswerable} while other models in the third block of Table~\ref{table:generation_performance} predict around or more than 20\%. As they are trained with the same dataset, we suspect this is due to their pretraining. It is also notable that predicting questions as \textit{unanswerable} is a difficult task for models and often affects the overall performance.

\begin{figure*}
    \centering
    \includegraphics[width=16cm]{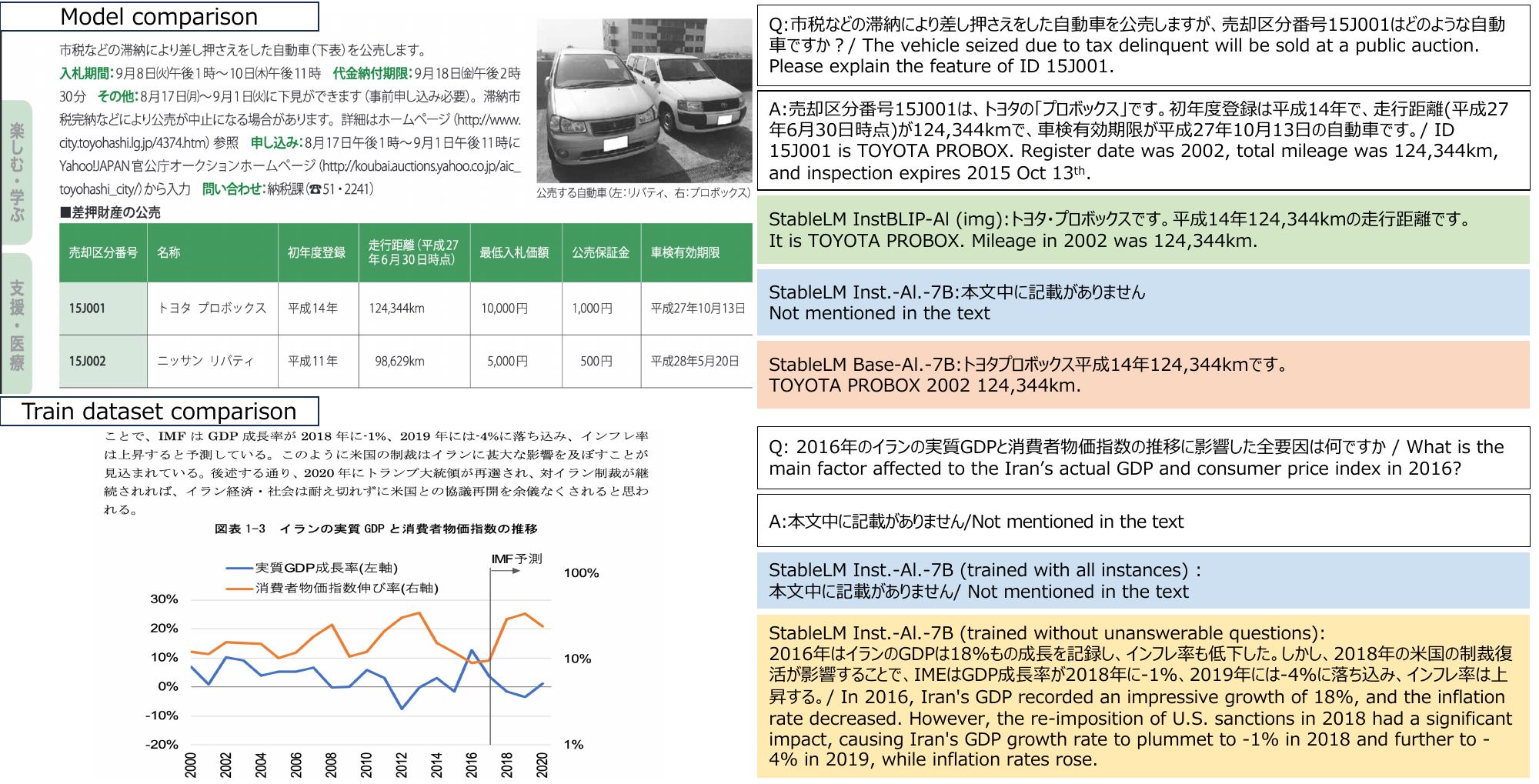}
    \caption{Qualitative analysis of open-ended question answering.}
    \label{fig:pdfqa_viz}
\end{figure*}

\paragraph{Multimodal model results.}
We present the results of models with multimodal inputs models of StableLM-InstructBLIP-Alpha in Table~\ref{table:multimodal_model}.
The model performances are enhanced especially when we use cropped images of referenced tables or figures (bbox).
We also notice that the model with black image inputs performs close to visual models to some extent, suggesting the effectiveness of textual inputs in our task.
We also carefully note that the max token length of StableLM-InstructBLIP-Alpha is 512, which can limit the textual understanding abilities of current multimodal models.

\paragraph{Token length dependency.}
We survey the token length effects on the performance.
For this purpose, we finetuned three models of different token lengths of rinna bi-4B-8k, e.g.,  2048, 4096, and 8192 respectively for both all training instances and without \textit{unanswerable} conditions. We present the performances in Table~\ref{table:different length}.
We notice the finetuning token length surely affects the final result, although we also notice finetuning models with long token length are much more computationally costly than those of short token length models, which can be the reason for the good performance of the rinna bi-4B-8k (8192 tokens) model in Table~\ref{table:generation_performance}.

\paragraph{Detailed Analyses of document-types.}
Table~\ref{table:detailed_file_type} presents the performance comparisons between each file type.
File types are classified into Japanese pamphlets such as public relations booklets or magazines, slides such as presentation materials, and report documents including figures and tables. We also prepare the out-of-domain test set of website scans where models still perform worse.

\paragraph{Qualitative Analysis.}
Figure~\ref{fig:pdfqa_viz} presents two example instances with the models' generations.
In the top example, we present the question, annotated answer, and generation from three models trained with questions including \textit{unanswerable}: StableLM InstructBLIP-Alpha (img), StableLM Instruct-Alpha-7B and StableLM Base-Alpha-7B respectively.
StableLM InstructBLIP-Alpha (img) can \textit{see} the alignments of the table via visual inputs and generate reasonable descriptions for the car. StableLM Base-Alpha also generates a similar answer while it cannot attribute the running mileage of 123,334 presented in the table. 
At the bottom of Figure~\ref{fig:pdfqa_viz}, we presented the comparison of two finetuned models that originate from the same pretrained model and are trained with and without \textit{unanswerable} questions. The all instances model accurately predicts that there are no answers written in the text while the model without \textit{unanswerable} questions falsely generates the open-ended answers, causing the phenomena known as hallucination.

\begin{table}[t]
\centering
\scriptsize\begin{tabular}{lcccc}
\hline 
\textbf{Model} &  \textbf{Human Evaluation $\uparrow$} \\
\hline
\multicolumn{2}{l}{\textit{Trained all and evaluated all.}} \\
PLaMo-13B                                 & 1.24 \\
StableLM Instruct-Alpha-7B                & 1.49 \\
\hline
StableLM InstructBLIP-Alpha (blank)       & 1.04 \\
StableLM InstructBLIP-Alpha (img)         & 1.25 \\
\hline
\end{tabular}
\caption{
Human evaluation on the sampled set.
}
\vspace{-1.7em}
\label{table:human_evaluation}
\end{table}

\paragraph{Human evaluation.}
Finally, we performed the human evaluation on the sampled results for some representative models.
We sampled 100 open-ended questions from the test set for this purpose.
We choose two text-input models of PLaMo-13B and StableLM Instruct-Alpha-7B. We also choose multimodal input models of StableLM InstructBLIP-Alpha with back and image inputs.
We ask an annotator to attach scores from 0 to 2 for two criteria: whether the generated answers include the annotated answer and whether the generated answers do not include wrong statements as questions' answers.
The results are presented in Table~\ref{table:human_evaluation}.
We notice that StableLM Instruct-Alpha-7B outperforms PLaMo-13B  and image model outperforms its blank image counterpart.

\section{Conclusion}
We introduced the JDocQA dataset concentrating the integration of both visual and linguistic cues in question answering in Japanese. We incorporated \textit{unanswerable} questions from given documents, which we confirmed is effective for harnessing the hallucinated generation to some extent in our experiments. 
Our detailed evaluations revealed the effectiveness of our dataset in a wide range of question categories from yes/no to open-ended and that the prediction of \textit{unanswerable} questions can be a clue to improve model performances, illustrating the effectiveness of the JDocQA dataset in realistic applications where multiple categories of questions are feasible and some extent of questions do not have explicit written answers in the documents.

\section{Acknowledgments}
This work was supported by JSPS Grant-in-Aid for Young Scientists (\#22K17983), 
 JSPS Fostering Joint International Research (A) (\#22KK0184) and by JST PRESTO (\#JPMJPR20C2).

\section*{Ethical Statement \& Limitation}
In data collection, we gather document PDF files and webpages from Japanese National Diet Library (NDL)'s digital collection, web archive projects (WARP), and websites of Japanese government ministries.
Administrative PDFs, documents, pamphlets, or websites published by local governments or universities are gathered through WARP.
We carefully avoid private documents and choose considerably public documents published by public or quasi-public sectors for the publicity of our dataset usage.
All of the documents and webpages are publicly available online and we follow our institutional rules to gather them.
We follow our institutional rules and also consult external advisors for data collection processes.

We assume our datasets are useful for both research and development of generative language models and their applications for Japanese document question answering. We also consider our dataset with \textit{unanswerable} questions can contribute to harnessing the hallucination problem of large language models.
However, this doesn't mean that the fintuned models with \textit{unanswerable} questions do not perform hallucinations at all.


\nocite{*}
\section{Bibliographical References}\label{sec:reference}

\bibliographystyle{lrec-coling2024-natbib}
\bibliography{lrec-coling2024-example}

\label{lr:ref}
\bibliographystylelanguageresource{lrec-coling2024-natbib}
\bibliographylanguageresource{languageresource}


\end{document}